\documentclass[sigconf]{acmart}

\usepackage{enumitem}
\usepackage{pifont}

\AtBeginDocument{%
  }

\copyrightyear{2026}
\acmYear{2026}
\setcopyright{cc}
\setcctype{by}
\acmConference[CIKM '26]{Proceedings of the 35th ACM International Conference on Information and Knowledge Management}{November 07--11, 2026}{Rome, Italy}
\acmBooktitle{Proceedings of the 35th ACM International Conference on Information and Knowledge Management (CIKM '26), November 07--11, 2026, Rome, Italy}
\acmDOI{10.1145/3799682.3839874}
\acmISBN{979-8-4007-2539-5/2026/11}

\begin{document}

\title{RecipeNet: A Hierarchical Transformer for Recipe Data}

\author{Pin-Yen Huang}
\affiliation{%
  \institution{Arizona State University}
  \city{Tempe}
  \state{Arizona}
  \country{USA}}
\email{pyhuang@asu.edu}

\author{Sachin Chhabra}
\affiliation{%
  \institution{Arizona State University}
  \city{Tempe}
  \state{Arizona}
  \country{USA}}
\email{sachin.chhabra@asu.edu}

\author{Prasanth Sai Gouripeddi}
\affiliation{%
  \institution{Arizona State University}
  \city{Tempe}
  \state{Arizona}
  \country{USA}}
\email{pgouripe@asu.edu}

\author{Abhinav Kumar}
\affiliation{%
  \institution{Applied Materials}
  \city{Santa Clara}
  \state{California}
  \country{USA}}
\email{abhinav_kumar@amat.com}

\author{Baoxin Li}
\affiliation{%
  \institution{University of Illinois Chicago}
  \city{Chicago}
  \state{Illinois}
  \country{USA}}
\email{baoxinli@uic.edu}

\renewcommand{\shortauthors}{Huang et al.}

\begin{abstract}

Recipe data arises in domains such as materials synthesis, pharmaceutical formulation, and industrial manufacturing, where procedures are represented as ordered sequences of steps containing heterogeneous structured fields. Existing tabular learning methods typically flatten this structure into fixed-schema representations, limiting their ability to capture hierarchical field interactions and procedural dependencies. We propose \textit{RecipeNet}, a hierarchical Transformer architecture that encodes field-level interactions within each step and sequential dependencies across steps through stacked Transformer encoders. Experiments on multiple recipe datasets and tasks demonstrate that RecipeNet consistently outperforms existing tabular models, highlighting the value of hierarchical and sequential modeling for recipe representation learning.

\end{abstract}



\maketitle

\begin{table}[t]
\small
\caption{Example of a manufacturing recipe sample.}
\label{tab:recipe_example}
\centering
\begin{tabular}{lccc}
\toprule
Attribute & Step 1: Mixing & Step 2: Heating & Step 3: Cooling \\
\midrule
Temperature & 25$^\circ$C & 80$^\circ$C & 30$^\circ$C \\
Pressure & --    & 2.5 bar & 1.0 bar \\
Duration & 5 min & 20 min & 10 min \\
Material A & 20 g & -- & -- \\
Material B & 10 g & -- & -- \\
\bottomrule
\end{tabular}
\end{table}

\section{Introduction}

Recipe data is a specialized form of tabular data that describes sequential procedures for producing a desired outcome. For example, in industrial manufacturing, recipes specify the sequence of operations, materials, and conditions required to produce a target product~\cite{brandl2006design, su2007control}.  Recipe data is also widely used across various domains, including cooking~\cite{yagcioglu2018recipeqa,lin2020recipe}, material synthesis~\cite{wang2022dataset}, pharmaceutical formulation~\cite{dong2024formulationai} and semiconductor manufacturing~\cite{chen2023improved,chen2025optimizing}. Despite differences across domains, these recipes share a common structure: each consists of heterogeneous fields organized into sequential procedural steps~\cite{bien2020recipenlg,wang2022dataset,dong2024formulationai}. An example of a manufacturing recipe is illustrated in Table~\ref{tab:recipe_example}.

Due to the varying field sets and numbers of procedural steps across recipes, existing tabular learning methods cannot be directly applied to recipe data. To use these methods, recipe data must first be converted into fixed-schema tabular representations through flattening and padding of the native recipe structure. These preprocessing steps often result in sparse representations and the loss of important structural relationships among fields and procedural steps~\cite{aggarwal2001surprising, hamilton2017representation}. More fundamentally, recipe data differs substantially from conventional tabular data in ways that make it difficult for existing tabular learning methods to model effectively~\cite{gorishniy2021revisiting,wu2020comprehensive}.

\noindent Specifically, recipe data presents three major challenges:

\begin{enumerate}[leftmargin=15pt]
\item \textbf{Variable schema.} Recipes contain varying field sets and numbers of procedural steps, whereas most tabular learning methods assume a fixed set of features across all samples.

\item \textbf{Hierarchical structure.} Recipes are naturally organized as procedural steps containing multiple fields. Effective modeling therefore requires capturing both intra-step field interactions and inter-step relationships, while conventional tabular methods typically operate on flat feature vectors.

\item \textbf{Sequential dependencies.} The order of procedural steps strongly influences the final outcome. However, many tabular learning methods are not designed to model sequential dependencies and generally treat feature ordering as irrelevant.

\end{enumerate}

\noindent To address these limitations, we propose RecipeNet, a hierarchical Transformer architecture for recipe data that preserves the native structure of procedural recipes. RecipeNet encodes heterogeneous field information within each procedural step and captures dependencies across steps through a second Transformer encoder. This design naturally accommodates variable recipe schemas while modeling both local field interactions and global procedural context. Experimental results show that RecipeNet consistently outperforms existing tabular learning methods across multiple recipe datasets and downstream prediction tasks, highlighting the effectiveness of hierarchical recipe modeling. The code is available at \url{https://github.com/pm25/recipenet}.

\section{Related Work}

\subsection{Fixed-Schema Tabular Learning}

Existing tabular learning methods are typically designed for fixed-schema tabular data, where all samples share the same set of features. Classical machine learning approaches, such as XGBoost~\cite{chen2016xgboost} and CatBoost~\cite{prokhorenkova2018catboost}, achieve strong performance on tabular data by capturing nonlinear relationships and interactions among features. More recently, deep tabular learning methods, such as TabNet~\cite{arik2021tabnet}, TabTransformer~\cite{huang2020tabtransformer}, FT-Transformer~\cite{gorishniy2021revisiting}, SAINT~\cite{somepalli2022saint}, and NODE~\cite{popov2019neural} have been proposed to learn feature representations from tabular inputs using neural architectures.

However, these methods are primarily designed for data with fixed feature layouts and do not explicitly model the variable schemas, hierarchical structure, and procedural dependencies characteristic of recipe data. Consequently, applying them to recipe datasets often requires flattening and padding to convert recipes into fixed-schema tabular representations, which can obscure the hierarchical and sequential structure inherent in recipe data.

\subsection{Structure-Aware Representation Learning}

Beyond conventional tabular learning, several model architectures have been proposed for modeling unordered sets or sequential data. Set-based architectures, including Deep Sets~\cite{zaheer2017deep} and Set Transformer~\cite{lee2019set}, learn permutation-invariant representations from unordered inputs. Sequential models such as recurrent neural networks (RNNs)~\cite{elman1990finding}, Long Short-Term Memory networks (LSTMs)~\cite{hochreiter1997long}, and Transformers~\cite{vaswani2017attention} have demonstrated strong performance in modeling temporal and sequential dependencies.

While these approaches effectively model either unordered sets or ordered sequences, recipe data combines both properties: each procedural step contains a variable set of heterogeneous fields, and the steps themselves form an ordered procedure. Existing methods are therefore not specifically designed to jointly model both aspects within a unified hierarchical framework.

\begin{figure}
    \centering
    \includegraphics[width=\linewidth]{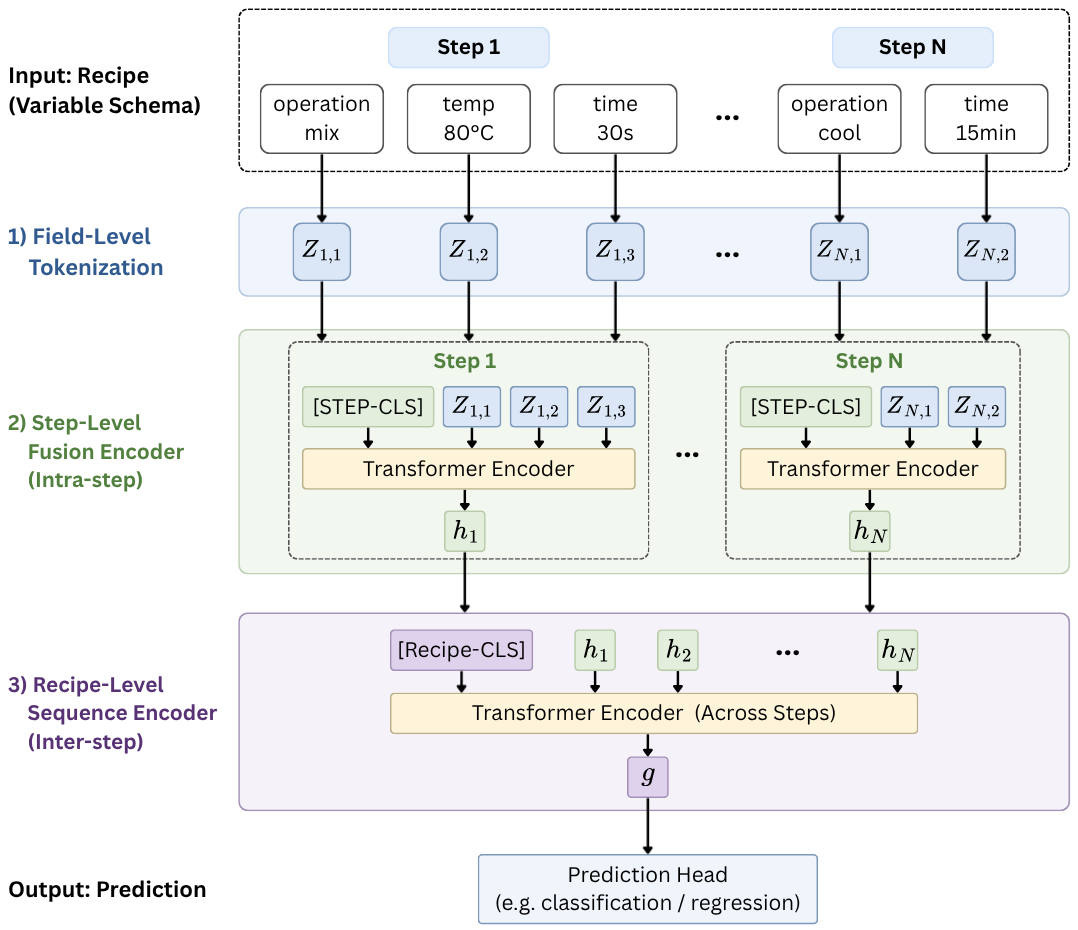}
    \caption{Overview of the proposed RecipeNet architecture.}
    \label{fig:recipenet}
\end{figure}

\section{Method}

\subsection{Overview}

We propose \textbf{RecipeNet}, a hierarchical Transformer architecture for recipe data. Unlike conventional tabular models that require preprocessing recipes into fixed-schema tabular representations, RecipeNet preserves the native hierarchical structure of recipes, where each recipe consists of an ordered sequence of procedural steps and each step contains a variable set of observed fields.  RecipeNet consists of three stages: (1) field-level tokenization (Section~\ref{method:field}), (2) step-level fusion (Section~\ref{method:step}), and (3) recipe-level sequence encoding (Section~\ref{method:seq}).

The architecture mirrors the hierarchical organization of recipe data. First, field-level tokenization maps heterogeneous numerical and categorical fields into a shared embedding space. A step-level Transformer then aggregates field embeddings within each procedural step to generate step representations. Finally, a recipe-level Transformer models dependencies across procedural steps to capture global procedural context and produce a recipe representation. This hierarchical design preserves the native structure of recipe data while naturally supporting variable schemas.

\subsection{Structured Recipe Representation}\label{method:repr}

A recipe is represented as an ordered sequence of procedural steps:
\[
R = \{S_1, S_2, \dots, S_N\}.
\]

Each step contains a variable number of observed fields. In RecipeNet, only observed fields are converted into tokens and included in the input sequence, allowing the model to naturally accommodate varying field sets and procedural structures.

\begin{table*}[t]
\centering
\caption{Performance comparison of tabular learning models across recipe datasets and downstream tasks.}
\label{exp:results}
\renewcommand{\arraystretch}{1.0}
\setlength{\tabcolsep}{9pt}
\begin{tabular}{lcccccc}
\hline
                          & \multicolumn{2}{c}{Solid-state reactions}                                                                                                                   & \multicolumn{2}{c}{Sol-gel precursor synthesis}                                                                                                             & \multicolumn{2}{c}{Solution synthesis}                                                                                                                      \\ \cline{2-7} 
Method                    & \begin{tabular}[c]{@{}c@{}}Next-step\\ Prediction \end{tabular} & \begin{tabular}[c]{@{}c@{}}Masked-step\\ Prediction \end{tabular} & \begin{tabular}[c]{@{}c@{}}Next-step\\ Prediction \end{tabular} & \begin{tabular}[c]{@{}c@{}}Masked-step\\ Prediction \end{tabular} & \begin{tabular}[c]{@{}c@{}}Next-step\\ Prediction \end{tabular} & \begin{tabular}[c]{@{}c@{}}Masked-step\\ Prediction \end{tabular} \\ \hline
XGBoost                   & 0.418$\pm$0.001                                                             & 0.974$\pm$0.000                                                               & 0.359$\pm$0.003                                                             & 0.911$\pm$0.003                                                               & 0.573$\pm$0.001                                                             & 0.882$\pm$0.001                                                               \\
CatBoost                  & 0.394$\pm$0.002                                                             & 0.993$\pm$0.001                                                               & 0.262$\pm$0.003                                                             & 0.981$\pm$0.004                                                               & 0.550$\pm$0.001                                                             & 0.980$\pm$0.001                                                               \\
TabNet                    & 0.367$\pm$0.008                                                             & 0.633$\pm$0.004                                                               & 0.363$\pm$0.008                                                             & 0.758$\pm$0.016                                                               & 0.501$\pm$0.005                                                             & 0.739$\pm$0.007                                                               \\
NODE                      & 0.167$\pm$0.000                                                             & 0.167$\pm$0.000                                                               & 0.167$\pm$0.000                                                             & 0.167$\pm$0.000                                                               & 0.167$\pm$0.000                                                             & 0.167$\pm$0.000                                                               \\
Transformer               & 0.439$\pm$0.010                                                             & 0.752$\pm$0.033                                                               & 0.331$\pm$0.026                                                             & 0.939$\pm$0.011                                                               & 0.546$\pm$0.004                                                             & 0.787$\pm$0.005                                                               \\
Set Transformer           & 0.173$\pm$0.006                                                             & 0.965$\pm$0.005                                                               & 0.238$\pm$0.001                                                             & 0.167$\pm$0.000                                                               & 0.506$\pm$0.001                                                             & 0.986$\pm$0.001                                                               \\
FT-Transformer            & 0.401$\pm$0.023                                                             & 0.595$\pm$0.061                                                               & 0.340$\pm$0.002                                                             & 0.842$\pm$0.023                                                               & 0.536$\pm$0.004                                                             & 0.771$\pm$0.000                                                               \\
TabTransformer            & 0.396$\pm$0.003                                                             & 0.682$\pm$0.017                                                               & 0.235$\pm$0.001                                                             & 0.911$\pm$0.017                                                               & 0.535$\pm$0.002                                                             & 0.908$\pm$0.019                                                               \\
\textbf{RecipeNet} (Ours) & \textbf{0.453$\pm$0.001}                                                    & \textbf{0.995$\pm$0.000}                                                      & \textbf{0.406$\pm$0.007}                                                    & \textbf{0.999$\pm$0.000}                                                      & \textbf{0.585$\pm$0.003}                                                    & \textbf{0.994$\pm$0.000}                                                      \\ \hline
\end{tabular}
\label{tab:results}
\end{table*}

\subsection{Field-Level Tokenization}\label{method:field}

RecipeNet maps observed field tokens into a shared latent space
of dimension \(d\). Numerical values are projected using a learned
linear layer:
\[
e^{(num)} = W_{num}x + b_{num},
\]
while categorical values are represented using embedding tables:
\[
e^{(cat)} = \mathrm{Embedding}_f(c).
\]

The resulting value embedding \(e^{(value)}\) is combined with a learned step-position embedding \(e^{(step)}\) and a learned field-identity embedding \(e^{(field)}\) to form the final token representation:
\[
z_{r,j} =
e^{(step)}_{r}
+
e^{(field)}_{r,j}
+
e^{(value)}_{r,j},
\]
where \(r\) denotes the step and \(j\) denotes the field token.

This tokenization scheme represents each field using its value, field identity, and procedural context, enabling RecipeNet to model heterogeneous field types without relying on fixed feature positions.

\subsection{Step Fusion Encoder}\label{method:step}

For each procedural step, RecipeNet applies a Transformer encoder over the observed field tokens within that step. A learnable \([\mathrm{STEP\mbox{-}CLS}]\) token is prepended to the sequence:
\[
h_n =
\mathrm{Transformer}_{step}
([\mathrm{STEP\mbox{-}CLS}], z_{n,1}, \dots, z_{n,m_n})_{[0]},
\]
where \(m_n\) denotes the number of observed field tokens in step \(n\). 

This step-level attention mechanism captures interactions among fields that jointly determine the behavior of a process step (e.g., pressure–temperature interactions), enabling RecipeNet to aggregate heterogeneous field into a contextualized step representation.

\subsection{Recipe-Level Sequence Encoder}\label{method:seq}

The sequence of step embeddings:
\[
H = \{h_1, h_2, \dots, h_N\},
\]
is processed using a second Transformer encoder operating over
procedural steps:
\[
g =
\mathrm{Transformer}_{recipe}
([\mathrm{RECIPE\mbox{-}CLS}], h_1, \dots, h_N)_{[0]}.
\]

The contextualized \([\mathrm{RECIPE\mbox{-}CLS}]\) token produces the final recipe representation \(g \in \mathbb{R}^{d}\). This recipe-level encoder captures long-range procedural dependencies across steps, enabling RecipeNet to model the overall procedural context of a recipe.

\subsection{Prediction Head}

The final recipe representation is passed to a task-specific prediction head for downstream tasks such as classification or regression:
\[
\hat{y} = Wg + b.
\]

RecipeNet naturally supports variable numbers of fields and procedural steps, heterogeneous numerical and categorical features, and hierarchical modeling of intra-step interactions and inter-step procedural dependencies. By preserving the hierarchical structure of recipe data, RecipeNet jointly captures local field interactions and global procedural context.

\section{Experiments}

\subsection{Experimental Setting}

\vspace{.2em}
\noindent\textbf{Datasets:}
We evaluate our method on three publicly available recipe datasets: the solid-state reaction dataset and the sol-gel precursor synthesis dataset from the Text-Mined Synthesis Project~\cite{kononova2019text}, and the solution synthesis dataset~\cite{wang2022dataset}.

\vspace{.2em}
\noindent\textbf{Downstream Tasks:}
Experiments were conducted on two downstream tasks: next-step prediction, and masked-step prediction.

\vspace{.2em}
\noindent\textbf{Evaluation Metrics:}
We report balanced accuracy for the tasks.

\vspace{.2em}
\noindent\textbf{Statistical Evaluation:}
Each experiment was conducted using the same three random seeds (0, 1, and 2). We report the mean and standard deviation across runs.

\vspace{.2em}
\noindent\textbf{Hyperparameters:}
All neural-based methods were trained using the same hyperparameter settings: learning rate \(1\times10^{-4}\), batch size \(32\), training iterations \(51{,}200\), and a cosine learning-rate scheduler~\cite{loshchilov2017sgdr}. Models were optimized using the AdamW optimizer~\cite{loshchilovdecoupled} with cross-entropy loss. Classical machine learning baselines used the default scikit-learn implementations~\cite{pedregosa2011scikit}.

\begin{figure*}[t]
    \centering 
    \includegraphics[width=1.0\linewidth]{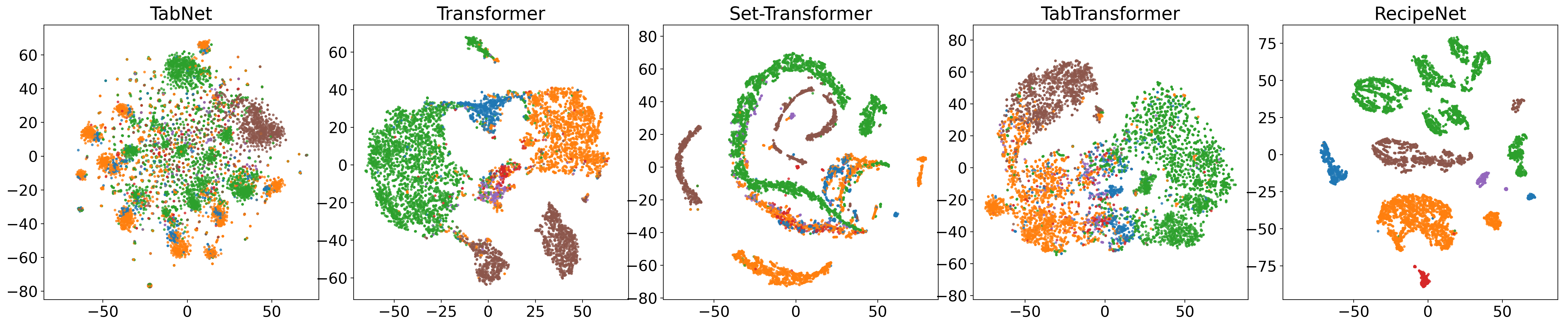} 
    \caption{Visualization of learned recipe representations using t-SNE. Different colors correspond to different target classes.} 
    \label{fig:vis} 
\end{figure*}

\begin{figure}[t]
    \centering
    \includegraphics[width=1.0\linewidth]{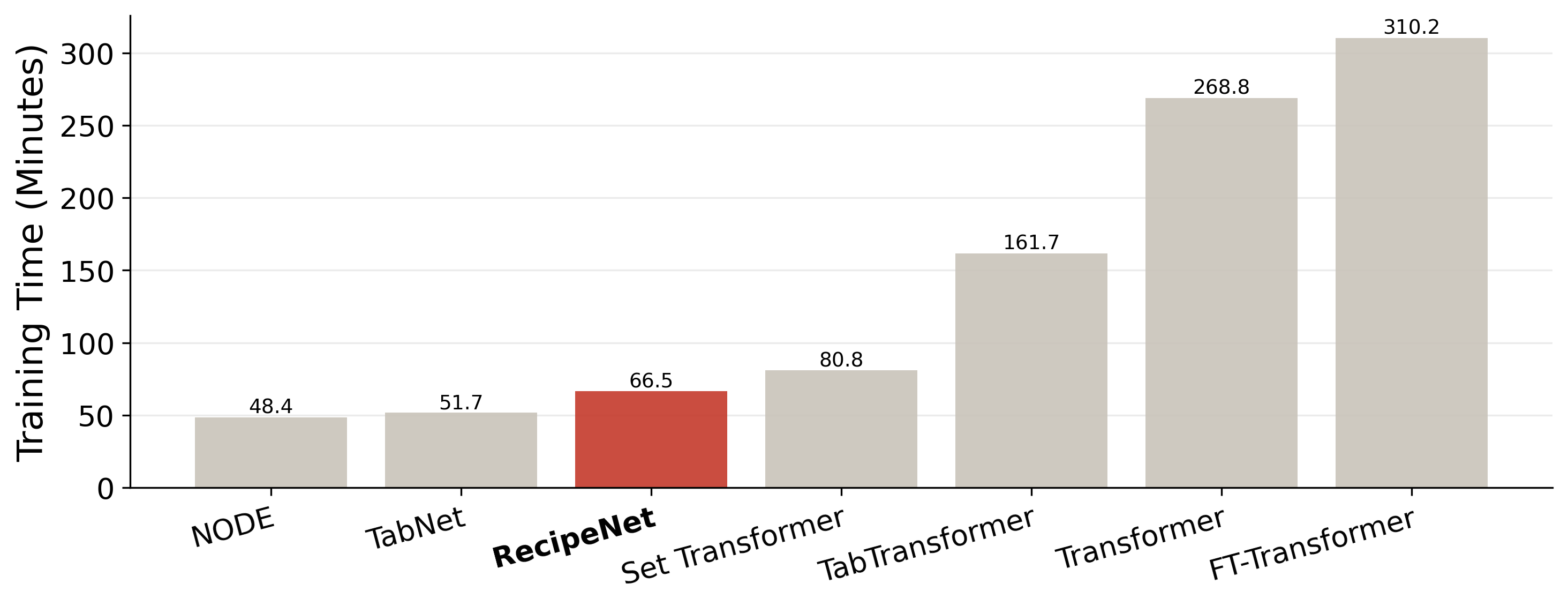}
    \caption{Comparison of total training time between RecipeNet and other neural tabular learning models.}
    \label{fig:speed} 
\end{figure}

\subsection{Analysis on Downstream Tasks}

To evaluate the effectiveness of RecipeNet, we compare its performance with existing tabular learning models on three recipe datasets across two downstream tasks: next-step prediction and masked-step prediction. The next-step prediction task aims to predict the type of the subsequent synthesis step given the preceding recipe context, while the masked-step prediction task requires recovering a masked step from the remaining recipe information. The results are summarized in Table~\ref{exp:results}.

RecipeNet consistently achieves the best performance across all datasets and tasks, outperforming tree-based methods, neural tabular models, and transformer-based architectures. The improvements are particularly evident in next-step prediction, where RecipeNet achieves the highest accuracy across all three synthesis domains. RecipeNet also attains near-perfect performance on masked-step prediction, consistently surpassing strong baselines such as XGBoost, CatBoost, and Set Transformer.

The strong performance of RecipeNet on both tasks suggests that explicitly modeling intra-step field interactions and inter-step procedural dependencies leads to more informative recipe representations. The consistent improvements across datasets and tasks demonstrate the effectiveness and robustness of the proposed hierarchical architecture for recipe data.

\subsection{Speed Analysis}

To evaluate the computational efficiency of RecipeNet, we compare its training time with other neural models on the Solid-state Reactions dataset. The reported times are averaged across multiple runs. As shown in Figure~\ref{fig:speed}, RecipeNet achieves the lowest training time among all transformer-based models. We attribute this efficiency to its hierarchical architecture and support for variable field tokens, which avoid unnecessary computation on missing recipe fields while compactly encoding the hierarchical structure of recipe data. As a result, RecipeNet reduces the computational overhead typically associated with transformer-based models.

Notably, RecipeNet reduces training time by approximately 18\% compared with the fastest competing transformer-based baseline while simultaneously achieving superior predictive performance. Some non-transformer models (NODE and TabNet) train faster, due to the simplicity of the model structure however they generally achieve lower predictive performance on the downstream tasks. These results suggest that RecipeNet achieves a favorable balance between computational efficiency and predictive accuracy, making it a practical approach for recipe data.

\subsection{Ablation Study}

To evaluate the contribution of each component in RecipeNet, we perform an ablation study by removing individual components and evaluating performance on the Sol-gel precursor synthesis dataset. The results are summarized in Table~\ref{exp:ablation}. The full model achieves the best performance on both next-step and masked-step prediction tasks, indicating that each component contributes to the overall effectiveness of RecipeNet.

The largest performance degradation is observed when removing the step encoder, followed by the recipe encoder. This suggests that modeling interactions among fields within a procedural step and dependencies across steps are both critical for recipe representation learning. Removing step-position embeddings or field-identity embeddings also reduces performance, demonstrating the importance of preserving procedural order and field-level semantic information. Finally, replacing the hierarchical architecture with a flattened representation leads to a performance drop, highlighting the benefit of explicitly modeling recipe data as a hierarchy of fields and procedural steps. The ablation results show that both the hierarchical structure and sequential modeling components contribute substantially to the performance of RecipeNet.

\begin{table}[]
\centering
\caption{Ablation study evaluating the contribution of each component in RecipeNet.}
\label{exp:ablation}
\renewcommand{\arraystretch}{1.0}
\setlength{\tabcolsep}{8pt}
\begin{tabular}{lcc}
\hline
Component                         & \begin{tabular}[c]{@{}c@{}}Next-step\\ Prediction\end{tabular} & \begin{tabular}[c]{@{}c@{}}Masked-step\\ Prediction\end{tabular} \\ \hline
RecipeNet (full model)            & \textbf{0.406}                                                          & \textbf{0.999}                                                            \\
w/o Hierarchy                     & 0.398                                                          & 0.995                                                            \\
w/o Step Encoder                  & 0.334                                                          & 0.973                                                            \\
w/o Recipe Encoder                & 0.351                                                          & 0.994                                                            \\
w/o Step Embedding                & 0.369                                                          & 0.997                                                            \\
w/o Field Embedding               & 0.391                                                          & 0.994                                                            \\ \hline
\end{tabular}
\end{table}

\subsection{Feature Visualization}

To better understand the representations learned by RecipeNet, we visualize recipe embeddings from the Sol-gel precursor synthesis dataset using t-SNE~\cite{van2008visualizing}. Figure~\ref{fig:vis} shows the projected embeddings, where each color denotes a target class. Compared with other baselines, RecipeNet produces more compact and well-separated clusters with reduced overlap between classes. In contrast, the baseline models exhibit greater inter-class mixing and more dispersed class distributions. The clear separation of target classes suggests that RecipeNet learns more discriminative representations, highlighting the benefits of hierarchical modeling for recipe data.

\section{Conclusion}

In this paper, we proposed RecipeNet, a hierarchical Transformer architecture designed for recipe data with variable schemas and sequential procedural structure. Unlike conventional tabular learning methods that flatten recipes into fixed representations, RecipeNet preserves both intra-step field interactions and inter-step procedural dependencies through hierarchical encoding. Experimental results across multiple recipe datasets and downstream tasks demonstrate the effectiveness of RecipeNet, highlighting the importance of explicitly modeling the hierarchical and sequential nature of recipe data for effective representation learning.

\begin{acks}
The authors gratefully acknowledge Applied Materials Inc. for its sponsorship and support of this work. The authors also acknowledge Research Computing at Arizona State University for providing computing resources that contributed to the research results reported in this paper. This research used the Sol supercomputer at Arizona State University~\cite{HPC:ASU23}.
\end{acks}

\bibliographystyle{ACM-Reference-Format}
\bibliography{sample-base}

\end{document}